\documentclass[10pt,twocolumn,letterpaper]{article}
\usepackage{float}
\usepackage{cvpr}
\usepackage{times}
\usepackage{epsfig}
\usepackage{graphicx}
\usepackage{amsmath}
\usepackage{amssymb}
\usepackage{etoolbox}
\patchcmd{\chapter}{plain}{empty}{}{}

\cvprfinalcopy

\usepackage[breaklinks=true,bookmarks=false]{hyperref}

\def\cvprPaperID{12} % *** Enter the CVPR Paper ID here

\begin{document}
\thispagestyle{empty}
%%%%%%%%% TITLE
\title{Apparent Age Estimation Using Ensemble of Deep Learning Models}

\author{Refik Can Mall{\i}\thanks{Both authors contributed equally to this work.}\\
\\
\\
{\tt\small }
\and
Mehmet Ayg\"un\footnotemark[1] \\ \\
Istanbul Technical University\\
Istanbul, Turkey\\
{\tt\small \{mallir,aygunme,ekenel\}@itu.edu.tr}
\and
Haz{\i}m Kemal Ekenel\\
{\tt\small }
}
\maketitle

%%%%%%%%% ABSTRACT
\begin{abstract}
\thispagestyle{empty}
In this paper, we address the problem of apparent age estimation. Different from estimating the real age of individuals, in which each face image has a single age label, in this problem, face images have multiple age labels, corresponding to the ages perceived by the annotators, when they look at these images. This provides an intriguing computer vision problem, since in generic image or object classification tasks, it is typical to have a single ground truth label per class. To account for multiple labels per image, instead of using average age of the annotated face image as the class label, we have grouped the face images that are within a specified age range. Using these age groups and their age-shifted groupings, we have trained an ensemble of deep learning models. Before feeding an input face image to a deep learning model, five facial landmark points are detected and used for 2-D alignment. We have employed and fine tuned convolutional neural networks (CNNs) that are based on VGG-16 ~\cite{Simonyan14c} architecture and pretrained on the IMDB-WIKI dataset ~\cite{Rothe-ICCVW-2015}. The outputs of these deep learning models are then combined to produce the final estimation. Proposed method achieves 0.3668 error in the final ChaLearn LAP 2016 challenge test set~\cite{LAP}.  

%Instead of using 101 outputs that represent each age as a class, we finetuned 3  CNN models using the ChaLearn LAP 2016 ~\cite{LAP} apparent age estimation’s competition train set images so that they have 34 outputs for generating overlap age group estimation. Then we combine these outputs of CNNs for predicting apparent age. Before training and testing images on CNNs, we detected and cropped faces with margin that includes all facial attributes. Facial landmark points are detected and used for 2-D alignment respect to 5 facial key points. Our method achieve 0.3668 error in final LAP challenge test set.  
\end{abstract}

\begin{figure}[!h]
\begin{center}
   \includegraphics[width=1\linewidth]{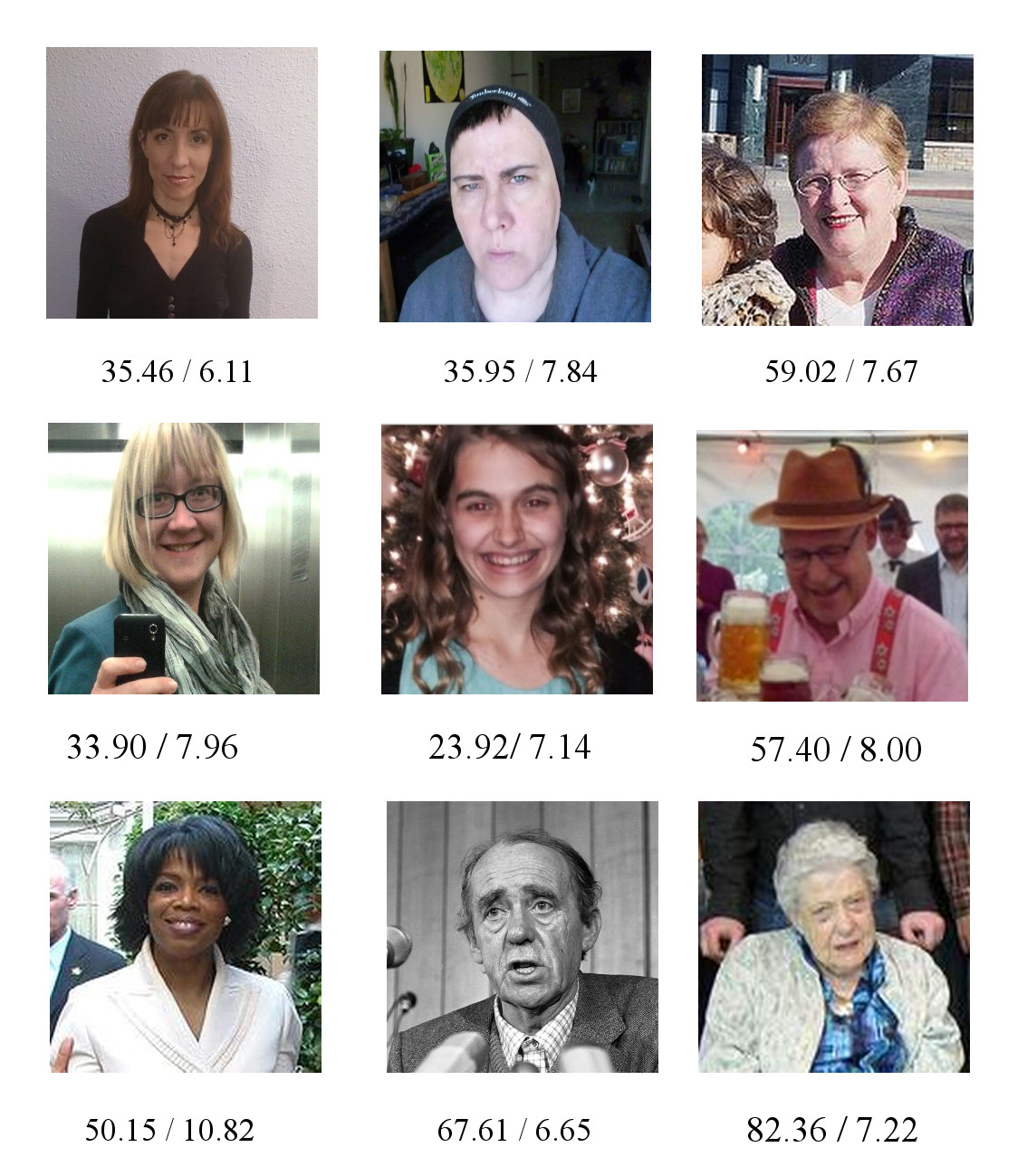}
\end{center}
   \caption{Sample images from the ChaLearn LAP 2016 apparent age estimation challenge dataset. Numbers below the images represent their average apparent age and standard deviations. The higher the standard deviation is, the harder it gets to predict apparent age of a person. Especially, in old age group, standard deviations are generally large, which makes it difficult to perform accurate age prediction from the images of old people. For instance, in LAP 2016 training set, there are 1095 images with standard deviations less than three, and only 31 of them belong to someone older than 40.}
\label{fig:samples}
\end{figure}

\section{Introduction}
\thispagestyle{empty}
Age estimation from face images is an interesting computer vision problem. It has various applications, ranging from customer relations to biometrics and entertainment. Although, there are various studies on real age estimation, ~\cite{X_geng,Fu_,Suo_} to name a few, problem of apparent age estimation from face images is a recently introduced topic, which has received significant attention ~\cite{LAP2015,ranjan2015unconstrained,kuang2015deeply,Rothe-ICCVW-2015,AgeNet}.

Besides the common difficulties in facial image processing and analysis, such as pose and illumination, the main challenges associated with age estimation have been the impact of ethnicity and subjective factors. The task of apparent age estimation alleviates the challenges posed by subjective factors. Since every person ages differently, real age may not be easy to deduce from face images. However, apparent age estimation is based on annotaters' perception of subjects' ages, who are displayed in the images. Therefore, the judgments for the age labes are expected to be based on visual appearance cues rather than personal characteristics. On the other hand, these perceived ages are subjective and they depend on the annotators, leading to result in multiple age labels for the same face image. This is a very intriguing problem, considering that for object classification problems, we normally have a single label per image. Sample images from the ChaLearn LAP 2016 dataset can be seen in Figure~\ref{fig:samples}.

In this study, we have proposed a novel approach for apparent age estimation, which addresses the problem of imprecise, uncertain multiple labels by grouping the face images that are within a specified age range, and by training an ensemble of deep learning models using these age groups and their age-shifted groupings. 

Convolutional Neural Networks (CNNs) have shown significant performance improvement in several computer vision problems, such as image classification ~\cite{ILSVRC15}, object detection~\cite{ILSVRC15}, image segmentation~\cite{crfasrnn_iccv2015}, and face recognition ~\cite{Parkhi15}. Moreover, all the best performing systems proposed in ChaLearn LAP 2015 ~\cite{LAP2015} were based on CNNs ~\cite{Rothe-ICCVW-2015,AgeNet,ucuncu}. Due to these reasons, we have opted for CNNs for the proposed system and have conducted a thorough study to efficiently transfer already existing models for the problem at hand. In addition, ensemble models are known to increase performance further. To benefit from this, we have generated different age groupings and train multiple CNN models. Finally, we have combined the outputs of these CNN models to produce the final estimation result.

The contributions of this study can be summarized as follows: (i) We proposed an apparent age estimation system that takes into account the imprecise, uncertain multiple labels available for the task. Instead of using avarage age labels as class labels, we have grouped the face images that are within a specified age range. An ensemble of CNNs have been trained using these age groups and their age-shifted groupings. (ii) We have conducted an extensive assessment about transferability of existing CNN models for apparent age estimation. (iii) We have analyzed the apparent age estimation problem in detail and pointed the challenges associated to it.

%In contrast to image classification ~\cite{imagenet_cvpr09} or object detection ~\cite{imagenet_cvpr09} datasets, there is only one large dataset named IMDB-WIKI~\cite{Rothe-ICCVW-2015} for age estimation. However, it is based real ages that crawled from web. Since there are not enough data for training CNNs from scratch, we started work on CNNs that have VGG-16 architecture which firstly pretrained on ImageNet then on IMDB-WIKI dataset. Moreover, apparent age classes are not easily separable because of that perception of age estimation differs from person to another. Therefore, we group adjacent apparent ages into 34 so that each of them consist of 3 ages for creating easily separable classes. We also started grouping from ages of 0,1 and 2 so that each groups have different overlapping ages and finetuned CNNs using these three groups separately so that each CNNs produce different but close predictions which simulates 3 different human. Finally, we ensemble these predictions for generating final age score.

The rest of the paper is organized as follows: In Section~2, a brief overview of related work is provided. In Section~3, the problem is stated and the challenges associated to it are pointed. The proposed method is explained in detail in Section~4. Experimental results are presented and discussed in Section~5. Finally, in Section~6, the paper is concluded with a brief summary and discussion. 

\section{Related Work}

%In this section, we first give a short overview about deep convolutional neural networks, then we will review CNN-based apparent age estimation methods.

Deep convolutional neural networks have performed impressively on very large-scale image recognition problems~\cite{Simonyan14c,resnet,alexnet}. Well known deep network architectures~\cite{Simonyan14c,GoogleNet} have been recently utilized in apparent age estimation studies. 
	
Deep EXpectation of Apparent Age From a Single Image (DEX) ~\cite{Rothe-ICCVW-2015}, which uses the CNN architecture presented in~\cite{Simonyan14c} was the winner of ChaLearn LAP 2015 apparent age estimation challenge~\cite{LAP}. In this method, the problem is approached as a classification problem with 101 age classes. They fine tuned VGG-16 model, which was pretrained on ImageNet~\cite{ILSVRC15}, using IMDB -WIKI~\cite{Rothe-ICCVW-2015} age dataset, that was created with the images crawled from the web. They splitted Chalearn LAP 2015~\cite{LAP2015} data to 20 different groups and trained 20 different models using these groups. Finally, they combined the output of each of the network using the weighted sum. The faces were not aligned in this study. They only performed face detection~\cite{Mathias} with different rotations and picked the one that had the best detection score. Also, they added 40\% margin to the faces before cropping. In case they cannot detect the face in the image, they used the entire image. They outperformed human performance and won the ChaLearn LAP 2015 apparent age estimation challenge~\cite{LAP}.
    
In contrast to the DEX method, AgeNet~\cite{AgeNet} approached to the problem as both classification and regression problem. They used large-scale deep convolutional neural networks~\cite{GoogleNet} for both tasks. Firstly, they normalized ages by dividing them with 100 and added a sigmoid layer as the output of regression model, so that the output is also normalized to [0,1]. In their second model, they trained their networks for classification task using Gaussian label distribution. While Euclidean loss was used for training phase of the regression model, cross-entropy loss was used for the classification model. All of the models were pretrained on face identification dataset using the CASIA-WebFace ~\cite{casia} and on age datasets ~\cite{cross,webface,morph}. Face detection and facial landmark localization were performed before face normalization. Two different face normalization methods, namely Exterior and Interior, were applied to the images. In total, eight different models were trained. Four of them were used for regression and four of them were used for classification. Two regression and two classification networks used the Exterior face template and the remaining four regression/classification networks used the Interior representation. Also, two different face crop sizes were used for training different networks. Finally, the output of each network were fused to produce the final estimation.

In contrast to DEX and AgeNet, in which the CNNs were used both for feature extraction and classification, Zhu \etal ~\cite{ucuncu} just used the CNNs as a feature extractor and employed additional classifiers, such as Support Vector Machines (SVM), Support Vector Regressors (SVR)~\cite{svr}, and Random Forests (RF)~\cite{randomforest} for apparent age estimation. Similar to the AgeNet method, they also trained their network first on the face identification dataset~\cite{casia} and then on the age datasets ~\cite{adiance,morph,lifespan,cross,webface}. They classified ages first into ten groups using the deep features that were extracted from CNN and were classified with SVM. After that, they used local age estimators, which used SVR and RF for the final predictions within each group.

\section{Problem Definition and Challenges}
\begin{figure}[h]
\begin{center}
   \includegraphics[width=1\linewidth]{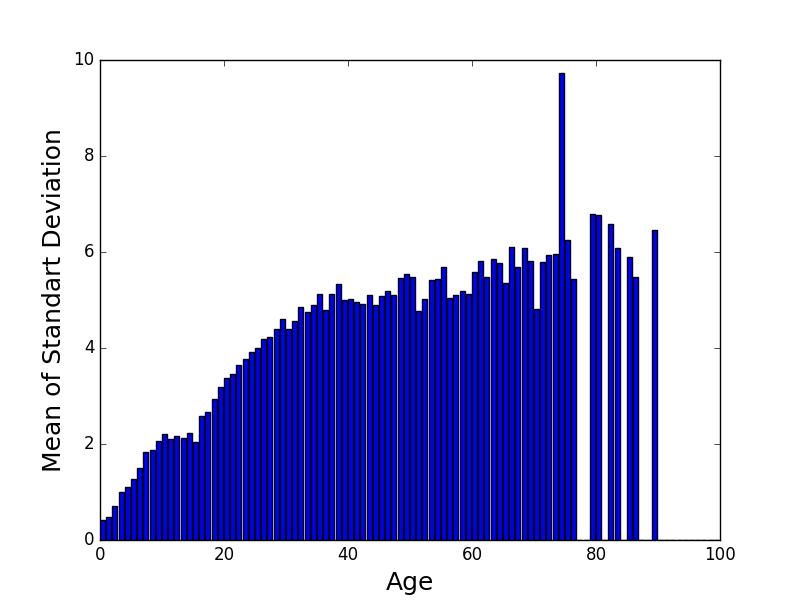}
    \caption{Mean value of standard deviation per age label increases when age labels increase.}
\label{fig:std}
\end{center}

\end{figure}

The ChaLearn LAP dataset~\cite{LAP} is based on apparent ages, which are annotated by more than 10 people for every image. In contrast to real age labels, apparent age labels are sensitive to environmental conditions, where the picture has taken and also changes with respect to annotators' perception. These reasons make the ground truth age labels uncertain. Standard deviation of the average age scores of the images are shown in Figure~\ref{fig:std}. It can be observed that mean value of standard deviation per age label increases, when the age increases. This indicates that estimating apparent age of the subjects, who are older than 30 is more challenging.

\begin{figure}[h]
\begin{center}
   \includegraphics[width=1\linewidth]{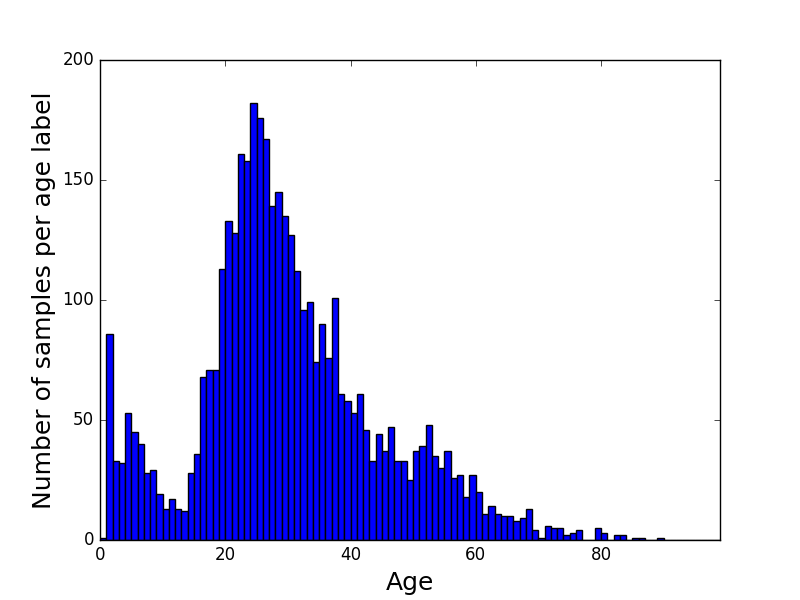}
    \caption{Age distribution of training set shows that dataset includes a lot of images between age of 20 and 40.}
\label{fig:dist}
\end{center}

\end{figure}

The competition dataset consists of 7591 images and it is divided into training, test, and validation sets. The number of images in each set is given in Table \ref{tab:datasetinfo}. Training and validation sets are provided with corresponding average age estimates by the annotators and their standard variations. Test set is spared for the final submission and the systems' performances are assessed on it. One of the observation that can clearly be made from Figure~\ref{fig:dist} is that the number of images provided for each age is not equal. Imbalanced data distribution is an important problem that should be handled by extracting representative features from the data without overfitting~\cite{inbalanced}.
    
\begin{table}[H]
\begin{center}
\begin{tabular}{|l|c|c|}
\hline
Part of Dataset & Number of Images \\
\hline\hline
Training & 4113 \\
Validation & 1500  \\
Test &1978 \\
Total &7591 \\
\hline
\end{tabular}
\end{center}
\caption{Number of images provided in the ChaLearn LAP 2016 dataset.}
\label{tab:datasetinfo}
\end{table}

\begin{figure}[h]
\begin{center}
   \includegraphics[width=0.8\linewidth]{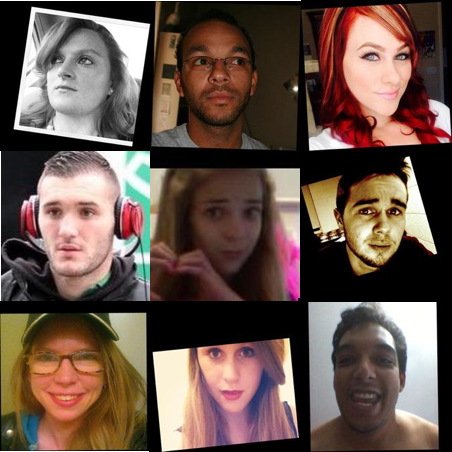}
    \caption{Sample face images from the training set that are labeled as 25 years old. Although the age labels are the same, facial attributes, image resolution, genders and ethnicities of people can be very different.}
\label{fig:25yas}
\end{center}

\end{figure}

Variation contained in the images, both in terms of image quality and facial appearance, is another challenge that prevents correct estimation of the apparent ages. Even though there are images that labeled with the same age, they could look significanly different from each other. Figure~\ref{fig:25yas} displays face images, which are labeled with an identical age, from different genders and ethnicities. Moreover, pictures are taken in different lighting conditions and some of them has photo effects on them. There are images that only a part of the face could be seen and sometimes they are covered with hair. Some of the people also have make-up on their faces or they wear accessories such as hats and glasses.

\section{Proposed Method}

\begin{figure*}[!t]
\begin{center}
	\includegraphics[width=1\linewidth]{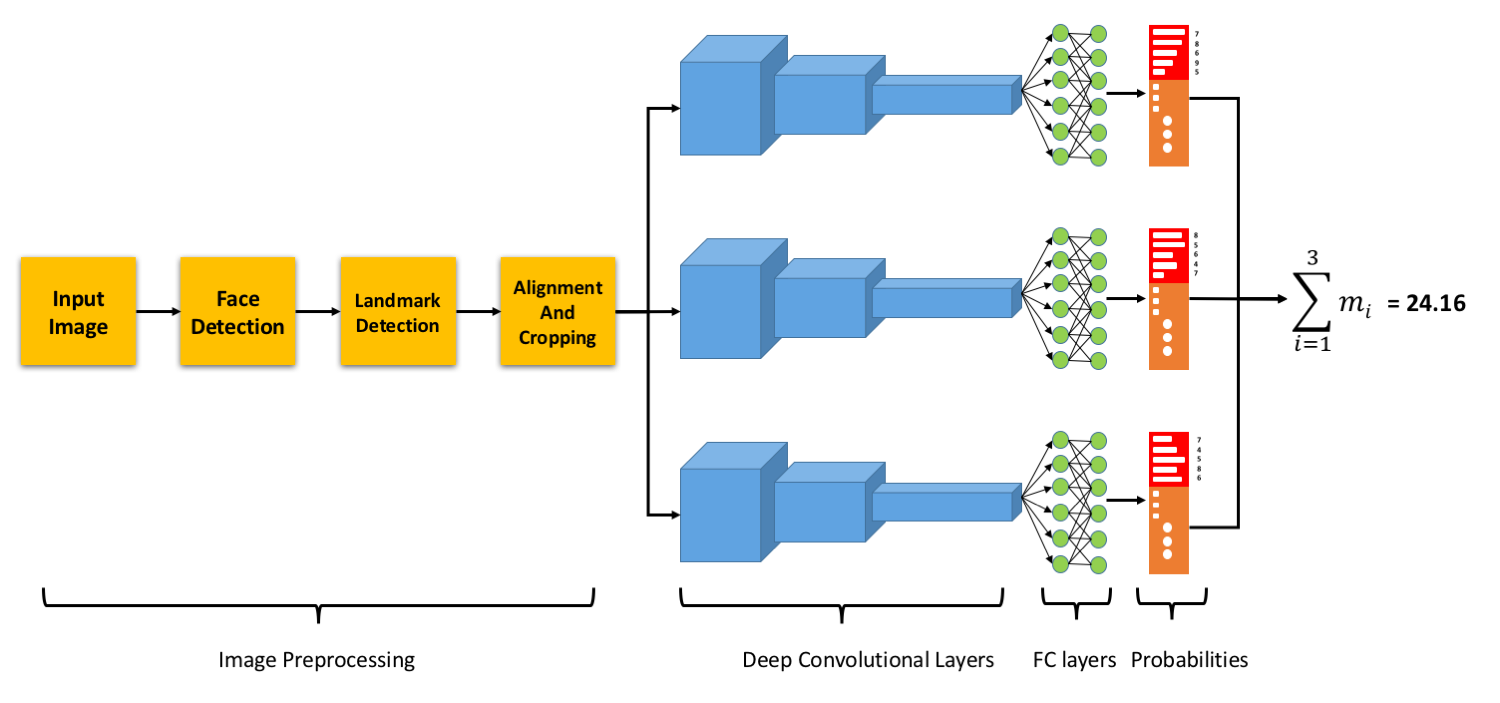}
\end{center}
   \caption{Overview of the proposed system. It consists of three main steps: In image preprocessing step, using Mathias \etal face detector, faces are found. Then five landmarks are located and faces are aligned with respect to these landmarks. Finally, aligned faces are cropped with 60\% additional margin. After preprocessing, three different VGG-16 model-based trained networks are fed with cropped faces. Using their softmax probabilities that are generated by the last fully connected layers (FC) of deep learning models ( $ m_{i} , i = 1,2,3 $), final score is obtained.}
\label{fig:overview}
\end{figure*}

Our proposed method follows the pipeline illustrated in Figure~\ref{fig:overview}. For system training, we have used the provided face images in the ChaLearn LAP 2016 apparent age estimation challenge dataset ~\cite{LAP}. The building blocks of the system are explained in the following subsections.

\subsection{Face Detection}

We have employed the Mathias \etal ~\cite{Mathias} face detector to obtain bounding boxes of faces in the images. Some images in the dataset contain more than one face. In this case, we have picked the faces with greater detection scores. To handle faces that appear at the corner or at the border of the images, we have added zero padding and created square sized images. Since, in 2\% of the images, the Mathias \etal face detector did not return any faces, we have utilized one more face detector, namely the one available in the dlib library ~\cite{dlib}, and run it on these images. In the end, only in 1.5\% of the images, no faces have been detected. The entire image has been fed to the CNN model in this case. In case of detection, the bounding boxes of faces have been enlarged by 30\% to benefit from more facial attributes for age estimation, i.e. hair and shape.

\subsection{Landmark Detection and Alignment}

\begin{figure}[!h]
\begin{center}
   \includegraphics[width=1\linewidth] {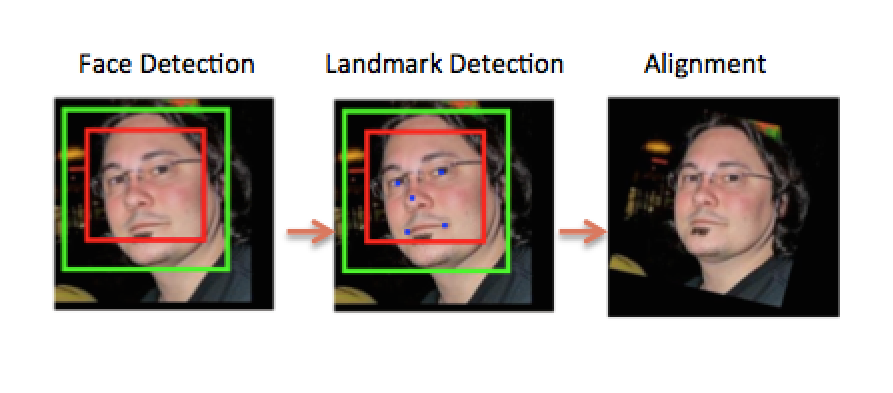}
    \caption{Overview of the face detection and alignment pipeline. Firstly, face is detected (red box) and 30\%  margin is added (green box). In the second step, five facial landmarks are localized and lastly the faces are aligned with respect to these landmarks.}
	\label{fig:alignment}
\end{center}
\end{figure}

For face alignment, we have first detected 68 facial landmarks using the approach presented in~\cite{onemilisecond}, which is available in the dlib ~\cite{dlib} library. After localizing 68 points, we have calculated five points (nose tip, eye centers, lip corners) and aligned the faces with respect to these points, as described in ~\cite{cinface}. In Figure~\ref{fig:alignment}, the red box shows the original bounding box from face detection and the green one represents the enlarged box.

\subsection{System Training}

In this subsection, we present the processes that have been performed for system training.
%The apparent age estimation results are obtained by using deep CNNs on detected and aligned face images. We have also augmented training dataset images to increase number of samples and reduce the effect of unbalanced class problem of dataset.

\subsubsection{Data Augmentation}

The competition dataset includes in total 4113 images for training and 1500 images for validation. Each age label is averaged over more than 10 different human voters. In Figure~\ref{fig:dist}, age distribution of training set is shown. It can be clearly seen that the distribution is not even. Moreover, the number of training samples is too low to train a deep CNN architecture. To tackle this problem, we have employed adaptive augmentation on the dataset, that increases the total number of images, while trying to make the age distribution of the set even. Our augmentation method includes random rotation, zooming, and color channel shifting. All images have been resized to $256 \times 256$ pixels resolution. We have also used five different cropped $224 \times 224$ pixels area of each image and combined them, while training the deep CNN models.

\subsubsection{Age - Shifted Grouping}

ChaLearn LAP 2016 dataset includes standard deviation of each ground truth age label. Average of standard deviations in the annotations of the dataset is 4.012. This causes a problem, since each face image has multiple age labels and the provided average face is not an exact label. To account for this problem, instead of using average age of the annotated face image as the class label, we have grouped the face images that are within a specified age range. Using these age groups and their age-shifted groupings, we have trained an ensemble of deep learning models.

\begin{figure}[H]
\begin{center}
   \includegraphics[width=1\linewidth]{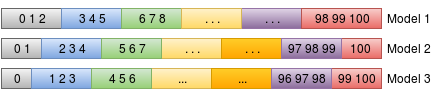}
    \caption{Shifted Age Groups}
    \label{fig:overlap}
\end{center}

\end{figure}

In contrast to grouping ages without changing boundaries of each group, which was proposed by Zhu \etal ~\cite{ucuncu}, we have implemented a grouping technique with a different perspective. Our proposed method consists of three different CNNs that group human ages into 34 classes. In addition to that, we have also shifted the age group indices by one for each deep model. In Figure~\ref{fig:overlap}, three different age groupings are illustrated. %Assigning different ages to shifted groups creates different perspective of estimating ages.

\subsubsection{Deep Learning with CNNs}

In this work, VGG-16 deep CNN models pretrained on the IMDB-WIKI dataset from~\cite{Rothe-ICCVW-2015} have been used for fine tuning on the challenge dataset. We have changed the last neural network layer of our networks to output 34 classes corresponding to different age groups. To prevent overfitting, we have changed the learning rate policy of models. They have been fine tuned for each layer separately. Learning rates of feature extraction layers are lowered, while fully connected layers' learning rates remain high. All models are trained with softmax loss function.

\subsubsection{Determination of Score Value}
	
To perform transition from age classification to age regression, i.e. to provide an age estimation output based on the CNN model classification outputs, classification results have been weighted summed with multiplication of their age indexes and probabilities. Unlike to the DEX method~\cite{Rothe-ICCVW-2015}, we have sorted probability vector output of the CNN models and selected predefined top $k$ elements and calculated weighted sum of them. This variable has also been tuned and we have selected the one that has produced the best score. At the last stage of age estimation, all three scores from three different models have summed up to produce the final output.

\section{Experiments}

Experiments have been conducted on the ChaLearn LAP 2016 validation set, which consists of 1500 images in total. As the error measure, we have used the same $\epsilon$-error function that was defined for the ChaLearn LAP challenge, which is given in Equation 1. Since LAP dataset has been annotated by multiple people,  ground truth value has been calculated as the average of votes, $\mu$, with standard deviation, $\sigma$. In error function $x$ represents the predicted age and the value of the error function can be between 1 (worst) and 0 (best). In the experiments, face detection and alignment procedures described in Sections 4.1 and 4.2 have been applied.

\begin{equation}
\epsilon = 1 - e ^ {-\dfrac{{(x-\mu)}^2}{2\sigma^2}}
\end{equation}

In the experiments, firstly, we have extracted deep features from FC-7 layer of VGG-FACE~\cite{Parkhi15} model and trained a 3-layer neural network that has two hidden layer using deep features. We have approached the problem as a multi-label classification problem and we have assigned age labels that are also in the boundaries of standard deviation, as true. This model performance has been found to be low with an error of 0.41, without aligning any detected faces. Then, we applied alignment on this setup and the alignment of the face images decreased our error rate to 0.39. Significant improvement is obtained after we used fine tuned VGG-16~\cite{Simonyan14c} models for testing our age-shifted grouping strategy. We have produced three different scores for each CNN, $m_{i}$, using sorted softmax probability vector, $p_{j}$, and corresponding ages, $\omega_{j}$. In the first experiment, we have used all of the probabilities and for the second one we have used only maximum probability for calculating the final score, which are formulated in Equations 2 and 3, respectively. 

\begin{equation}
m_{i} = \sum\limits_{j=1}^{34} p_{j} \omega_{j} 
\end{equation}

\begin{equation}
m_{i} =  p_{1} \omega_{1} 
\end{equation}

Lastly, we have used the five maximum probabilities to produce the final score of the CNN, as formulated in Equation 4, which has attained the best accuracy in the validation set.
    
\begin{equation}
m_{i} = \sum\limits_{j=1}^{5} p_{j} \omega_{j} 
\end{equation}

Finally, we have combined these $m_{i}$ values that represent the outputs of CNNs to estimate the final score of the system using Equation \ref{eq:final}. Since all $m_{i}$ values are in range of 0-34, they are summed up to produce the final prediction. All experiments are implemented with Caffe Deep Learning Framework~\cite{caffe} on NVIDIA TITAN X GPU. Training of each deep model took approximately 5 hours. The experimental results of these different cases are listed in Table~\ref{tab:tabloresult}.

\begin{equation}
prediction  = \sum\limits_{i=1}^{3} m_{i} 
\label{eq:final}
\end{equation}

\begin{table}[H]
\begin{center}
\begin{tabular}{|l|c|}
\hline
Method & Error \\
\hline\hline
VGG-FACE + w/o Alignment + Neural Network & 0.4128 \\
VGG-FACE + Alignment + Neural Network & 0.3906 \\
3 CNNs + Alignment + full weighted sum. & 0.2907 \\
3 CNNs + Alignment + max. 1 weighted sum. & 0.3000\\
3 CNNs + Alignment + max. 5 weighted sum. & 0.2897\\
\hline
\end{tabular}
\end{center}
\caption{Experimental Results}
\label{tab:tabloresult}
\end{table}

\begin{table}[H]
\begin{center}
\begin{tabular}{|l|c|c|}
\hline
Rank & Team & Test Error \\
\hline\hline
1 & OrangeLabs & 0.2411 \\
2 &palm\_seu & 0.3214 \\
3 &cmp+ETH & 0.3361\\
4 &WYU\_CVL & 0.3405\\
5 & \textbf {ITU\_SiMiT (Ours)} & 0.3668\\
6 &Bogazici & 0.3740\\
7 &MIPAL\_SNU & 0.4569\\
8 &DeepAge & 0.4573\\
\hline
\end{tabular}
\end{center}
\caption{ChaLearn LAP 2016 final ranking on the test set. Out of 84 registered participants, there were only eight submissions.}
\label{tab:tablochalearn}
\end{table}

\begin{figure}
\begin{center}
   \includegraphics[width=1\linewidth]{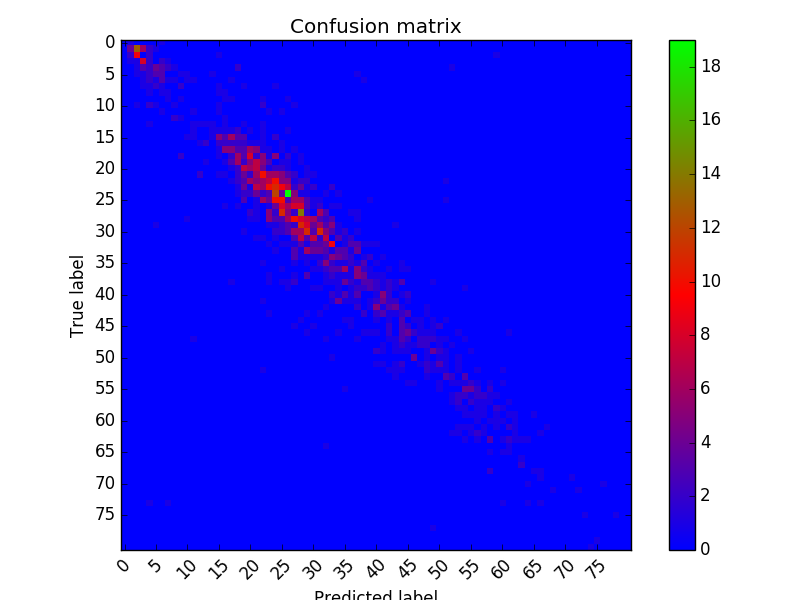}
    \caption{Confusion matrix obtained on the validation set (Best displayed in color).}
\end{center}
\label{fig:comatrix}
\end{figure}

\section{Conclusions}

In this study, we have built an ensemble of deep learning models for apparent age estimation. Together with age groupings, this ensemble model accounts for imprecise, uncertain apparent age labels of the face images. In the proposed system, we have employed five-point face alignment. To equalize age distribution in the dataset, we have performed an adaptive data augmentation. We have utilized different fusion schemes and with these enhancements, we have attained an error rate on the validation set, which is on par with the last year's best performing system. However, the error obtained on the test set, 0.3668, is relatively higher than the one obtained on the validation set. As a future work, we will investigate the reasons for this performance gap. Furthermore, we plan to develop a training scheme that takes into account the standard deviations of the labels.

\section*{Acknowledgements}

This work was supported by TUBITAK project no. 113E067 and by a Marie Curie FP7 Integration Grant within the 7th EU Framework Programme.

{\small
\bibliographystyle{ieee}
\bibliography{egbib}

\begin{thebibliography}{10}\itemsep=-1pt

\bibitem{randomforest}
L.~Breiman.
\newblock Random forests.
\newblock {\em Machine learning}, 45(1):5--32, 2001.

\bibitem{cross}
B.-C. Chen, C.-S. Chen, and W.~H. Hsu.
\newblock Cross-age reference coding for age-invariant face recognition and
  retrieval.
\newblock In {\em Computer Vision--ECCV 2014}, pages 768--783. Springer, 2014.

\bibitem{adiance}
E.~Eidinger, R.~Enbar, and T.~Hassner.
\newblock Age and gender estimation of unfiltered faces.
\newblock {\em Information Forensics and Security, IEEE Transactions on},
  9(12):2170--2179, 2014.

\bibitem{LAP2015}
S.~Escalera, J.~Fabian, P.~Pardo, X.~Baro, J.~Gonzalez, H.~J. Escalante,
  D.~Misevic, U.~Steiner, and I.~Guyon.
\newblock Chalearn looking at people 2015: Apparent age and cultural event
  recognition datasets and results.
\newblock In {\em The IEEE International Conference on Computer Vision (ICCV)
  Workshops}, December 2015.

\bibitem{LAP}
S.~Escalera, M.~Torres, B.~Martínez, X.~Baró, H.~J. Escalante, I.~Guyon,
  G.~Tzimiropoulos, C.~Corneanu, M.~Oliu, M.~A. Bagheri, and M.~Valstar.
\newblock Chalearn looking at people and faces of the world: Face analysis
  workshop and challenge 2016.
\newblock In {\em CVPR workshops}, 2016.

\bibitem{Fu_}
Y.~Fu and T.~S. Huang.
\newblock Human age estimation with regression on discriminative aging
  manifold.
\newblock 2008.

\bibitem{X_geng}
X.~Geng, Z.~H. Zhou, and K.~Smith-Miles.
\newblock Automatic age estimation based on facial aging pattern.
\newblock In {\em IEEE Transactions on Pattern Analysis and Machine
  Intelligence}, 2007.

\bibitem{inbalanced}
H.~He and E.~A. Garcia.
\newblock Learning from imbalanced data.
\newblock {\em Knowledge and Data Engineering, IEEE Transactions on},
  21(9):1263--1284, 2009.

\bibitem{resnet}
K.~He, X.~Zhang, S.~Ren, and J.~Sun.
\newblock Deep residual learning for image recognition.
\newblock {\em CoRR}, abs/1512.03385, 2015.

\bibitem{caffe}
Y.~Jia, E.~Shelhamer, J.~Donahue, S.~Karayev, J.~Long, R.~Girshick,
  S.~Guadarrama, and T.~Darrell.
\newblock Caffe: Convolutional architecture for fast feature embedding.
\newblock {\em arXiv preprint arXiv:1408.5093}, 2014.

\bibitem{onemilisecond}
V.~Kazemi and J.~Sullivan.
\newblock One millisecond face alignment with an ensemble of regression trees.
\newblock In {\em Proceedings of the IEEE Conference on Computer Vision and
  Pattern Recognition}, pages 1867--1874, 2014.

\bibitem{dlib}
D.~E. King.
\newblock Dlib-ml: A machine learning toolkit.
\newblock {\em Journal of Machine Learning Research}, 10:1755--1758, 2009.

\bibitem{alexnet}
A.~Krizhevsky, I.~Sutskever, and G.~E. Hinton.
\newblock Image{N}et classification with deep convolutional neural networks.
\newblock In {\em Advances in neural information processing systems}, pages
  1097--1105, 2012.

\bibitem{kuang2015deeply}
Z.~Kuang, C.~Huang, and W.~Zhang.
\newblock Deeply learned rich coding for cross-dataset facial age estimation.
\newblock In {\em Proceedings of the IEEE International Conference on Computer
  Vision Workshops}, pages 96--101, 2015.

\bibitem{AgeNet}
X.~Liu, S.~Li, M.~Kan, J.~Zhang, S.~Wu, W.~Liu, H.~Han, S.~Shan, and X.~Chen.
\newblock Agenet: Deeply learned regressor and classifier for robust apparent
  age estimation.
\newblock In {\em The IEEE International Conference on Computer Vision (ICCV)
  Workshops}, December 2015.

\bibitem{Mathias}
M.~Mathias, R.~Benenson, M.~Pedersoli, and L.~{Van Gool}.
\newblock Face detection without bells and whistles.
\newblock In {\em ECCV}, 2014.

\bibitem{lifespan}
M.~Minear and D.~C. Park.
\newblock A lifespan database of adult facial stimuli.
\newblock {\em Behavior Research Methods, Instruments, \& Computers},
  36(4):630--633, 2004.

\bibitem{webface}
B.~Ni, Z.~Song, and S.~Yan.
\newblock Web image and video mining towards universal and robust age
  estimator.
\newblock {\em Multimedia, IEEE Transactions on}, 13(6):1217--1229, 2011.

\bibitem{Parkhi15}
O.~M. Parkhi, A.~Vedaldi, and A.~Zisserman.
\newblock Deep face recognition.
\newblock In {\em British Machine Vision Conference}, 2015.

\bibitem{ranjan2015unconstrained}
R.~Ranjan, S.~Zhou, J.~Chen, A.~Kumar, A.~Alavi, V.~Patel, and R.~Chellappa.
\newblock Unconstrained age estimation with deep convolutional neural networks.
\newblock In {\em Proceedings of the IEEE International Conference on Computer
  Vision Workshops}, pages 109--117, 2015.

\bibitem{morph}
K.~Ricanek~Jr and T.~Tesafaye.
\newblock Morph: A longitudinal image database of normal adult age-progression.
\newblock In {\em Automatic Face and Gesture Recognition, 2006. FGR 2006. 7th
  International Conference on}, pages 341--345. IEEE, 2006.

\bibitem{Rothe-ICCVW-2015}
R.~Rothe, R.~Timofte, and L.~V. Gool.
\newblock {DEX}: Deep {EX}pectation of apparent age from a single image.
\newblock In {\em ICCV, Cha{L}earn Looking at People workshop}, December 2015.

\bibitem{ILSVRC15}
O.~Russakovsky, J.~Deng, H.~Su, J.~Krause, S.~Satheesh, S.~Ma, Z.~Huang,
  A.~Karpathy, A.~Khosla, M.~Bernstein, A.~C. Berg, and L.~Fei-Fei.
\newblock {Image{N}et Large Scale Visual Recognition Challenge}.
\newblock {\em International Journal of Computer Vision (IJCV)},
  115(3):211--252, 2015.

\bibitem{Simonyan14c}
K.~Simonyan and A.~Zisserman.
\newblock Very deep convolutional networks for large-scale image recognition.
\newblock {\em CoRR}, abs/1409.1556, 2014.

\bibitem{svr}
A.~Smola and V.~Vapnik.
\newblock Support vector regression machines.
\newblock {\em Advances in neural information processing systems}, 9:155--161,
  1997.

\bibitem{Suo_}
J.~Suo, T.~Wu, S.~Zhu, and S.~Shan.
\newblock Design sparse features for age estimation using hierarchical face
  model.
\newblock In {\em IEEE International Conference on Automatic Face and Gesture
  Recognition}, 2008.

\bibitem{GoogleNet}
C.~Szegedy, W.~Liu, Y.~Jia, P.~Sermanet, S.~Reed, D.~Anguelov, D.~Erhan,
  V.~Vanhoucke, and A.~Rabinovich.
\newblock Going deeper with convolutions.
\newblock {\em CoRR}, abs/1409.4842, 2014.

\bibitem{cinface}
X.~Wu, R.~He, and Z.~Sun.
\newblock A lightened {CNN} for deep face representation.
\newblock {\em arXiv preprint arXiv:1511.02683}, 2015.

\bibitem{casia}
D.~Yi, Z.~Lei, S.~Liao, and S.~Z. Li.
\newblock Learning face representation from scratch.
\newblock {\em arXiv preprint arXiv:1411.7923}, 2014.

\bibitem{crfasrnn_iccv2015}
S.~Zheng, S.~Jayasumana, B.~Romera-Paredes, V.~Vineet, Z.~Su, D.~Du, C.~Huang,
  and P.~Torr.
\newblock Conditional random fields as recurrent neural networks.
\newblock In {\em International Conference on Computer Vision (ICCV)}, 2015.

\bibitem{ucuncu}
Y.~Zhu, Y.~Li, G.~Mu, and G.~Guo.
\newblock A study on apparent age estimation.
\newblock In {\em The IEEE International Conference on Computer Vision (ICCV)
  Workshops}, December 2015.

\end{thebibliography}
}

\begin{figure*}
\begin{center}
	\includegraphics[width=1\linewidth]{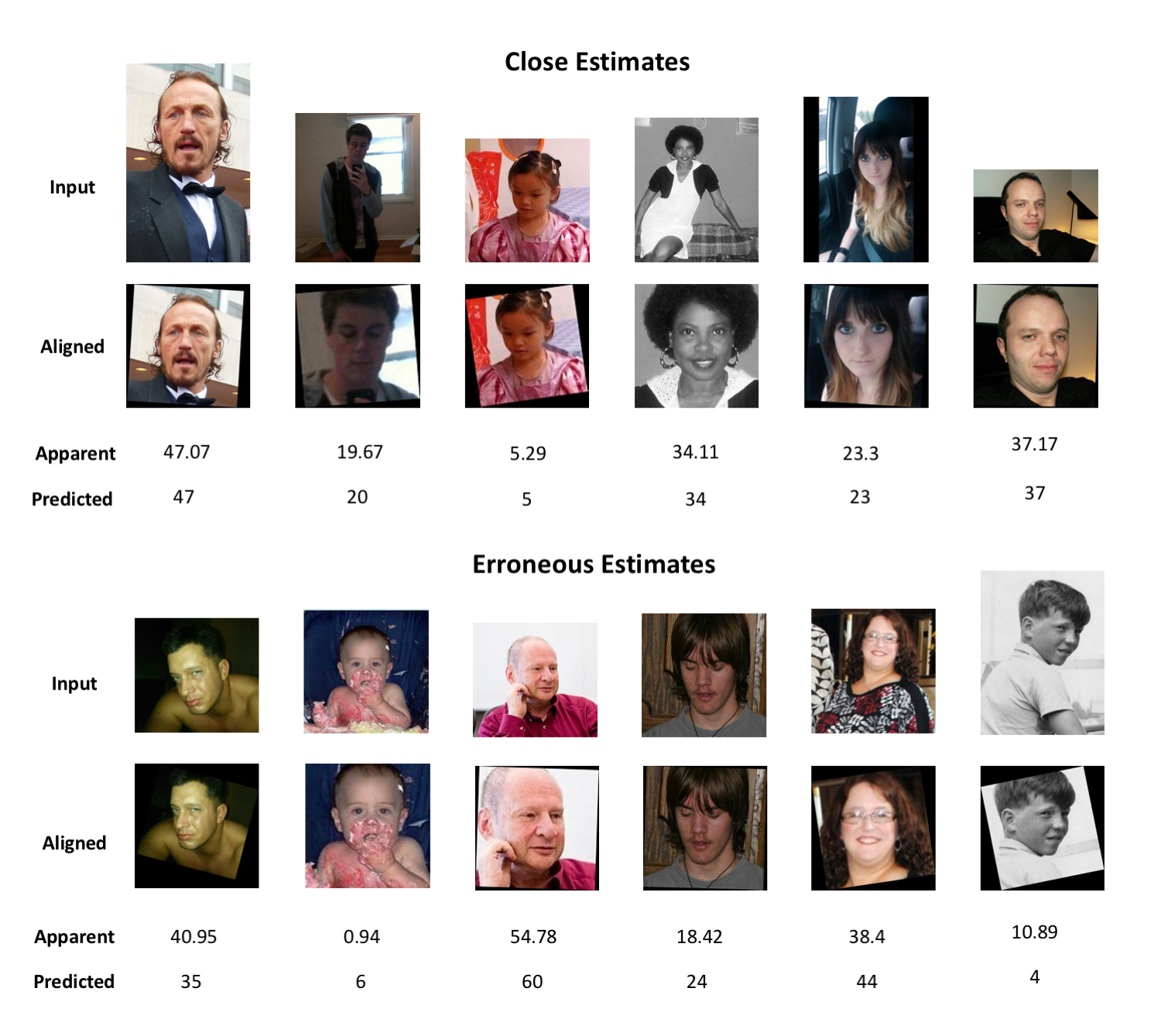}
\end{center}
   \caption{Close and erroneous estimates from sample images. }
\label{fig:badgood}
\end{figure*}

\end{document}